\documentclass[acmsmall]{acmart}
\usepackage{hyperref}
\usepackage{xcolor}
\AtBeginDocument{%
  }

\setcopyright{cc}
\setcctype{by}
\acmJournal{PACMCGIT}
\acmYear{2026} \acmVolume{9} \acmNumber{3} \acmArticle{42}
\acmMonth{7} \acmDOI{10.1145/3816072}

\begin{document}

\title{Echoes of the Prior: A Computational Phenomenology of Forgetting}

\author{Gege Gao}
\email{gege.gao@uni-tuebingen.de}
\affiliation{%
  \institution{Eberhard Karl University of Tübingen}
  \city{Tübingen}
  \state{Baden-Württemberg}
  \country{Germany}
}

\author{Bernhard Schölkopf}
\authornote{Bernhard Schölkopf and Andreas Geiger jointly advised this work.}
\email{bs@tuebingen.mpg.de}
\affiliation{%
  \institution{Max Planck Institute for Intelligent Systems}
  \city{Tübingen}
  \state{Baden-Württemberg}
  \country{Germany}
}

\author{Andreas Geiger}
\authornotemark[1]
\email{a.geiger@uni-tuebingen.de}
\affiliation{
  \institution{Eberhard Karl University of Tübingen, Tübingen AI Center}
  \city{Tübingen}
  \state{Baden-Württemberg}
  \country{Germany}
}


 
\begin{teaserfigure}
    \includegraphics[width=\textwidth]{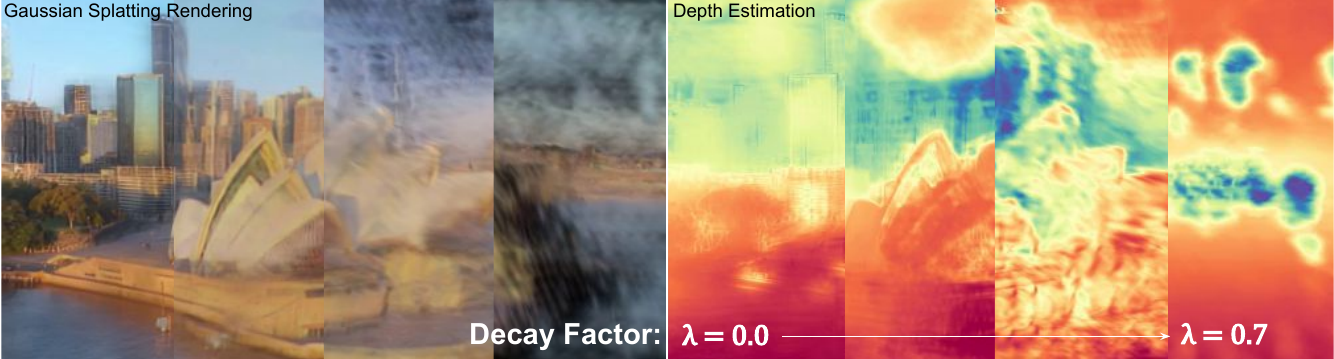}
    \caption{
        \textbf{\textit{Echoes of the Prior}}. A computational simulation of the fragility of memory under entropic loss. 
    }
    \Description{.}
    \label{fig:teaser}
\end{teaserfigure}

\begin{abstract}
    Memory is not merely the storage of data; it is the scaffolding of reality. When biological memory fades, the world does not simply turn black; it regresses into an unrecognizable chaos. \textit{Echoes of the Prior} is an interactive installation that attempts to visualize this subjective phenomenology of \textit{forgetting}. By inducing controlled synaptic decay within a Feed-Forward 3D Reconstruction model, we create an artistic analogy for the erosion of the brain's predictive priors. We position the Neural Network not as a tool for engineering, but as a cognitive proxy - a \textit{silicon brain} whose structural degeneration evokes the disorienting, poetic, and terrifying experience of losing one's grip on the world. Ultimately, we offer this framework as a catalyst, inviting the wider community to explore the uncharted potential of neuromorphic aesthetics in visualizing the fragility of intelligence. Interactive demo: \url{https://decart-4d.github.io/}. 
\end{abstract}




\maketitle

\section{Introduction: The Invisible Process}

``What I cannot create, I do not understand'' is a famous quote by physicist Richard Feynman. While neuroscience can explain the \textit{mechanism} of Alzheimer's -- the plaques, the tangles, the synaptic pruning -- it cannot show us the experience. We know that the patient forgets, but we do not know \textit{what the world looks like} to them as it fades. 

Modern Vision Transformers share structural principles with biological vision -- distributed representation, hierarchical feature extraction, and predictive coding~\cite{yamins2016using}. While these correspondences are productive analogies rather than mechanistic equivalences~\cite{bowers2023deep, lindsay2021convolutional}, they invite a speculative proposition: if these models can approximate aspects of human visual cognition, then their pathology may serve as an \textit{artistic analogy} for the pathology of the mind.

\textit{Echoes of the Prior} operates on this speculative premise. We treat a state-of-the-art 3D Gaussian Splatting (3DGS) model not as a renderer, but as a proxy observer -- a \textit{silicon brain} whose components we map, by analogy, onto cognitive functions. By surgically introducing noise into its semantic priors (analogous to \textit{long-term memory}) and image encoder (analogous to the \textit{sensory cortex}), we force the machine to reconstruct reality through a damaged architecture. The resulting imagery -- melting sky, ghostly architectures, and dissolving forms -- serves as a visual translation of fading mind, making the invisible internal experience of forgetting visible, visceral, and shared.

We title our installation \textit{\textbf{DecArt}} (a portmanteau of \textit{Decay} and \textit{Art}). Pronounced like the French philosopher \textit{Descartes}, we elaborate on the philosophical implications of this choice in Sec.~\ref{sec:name}. 

\begin{figure}[t]
    \centering
    \includegraphics[width=0.99\linewidth]{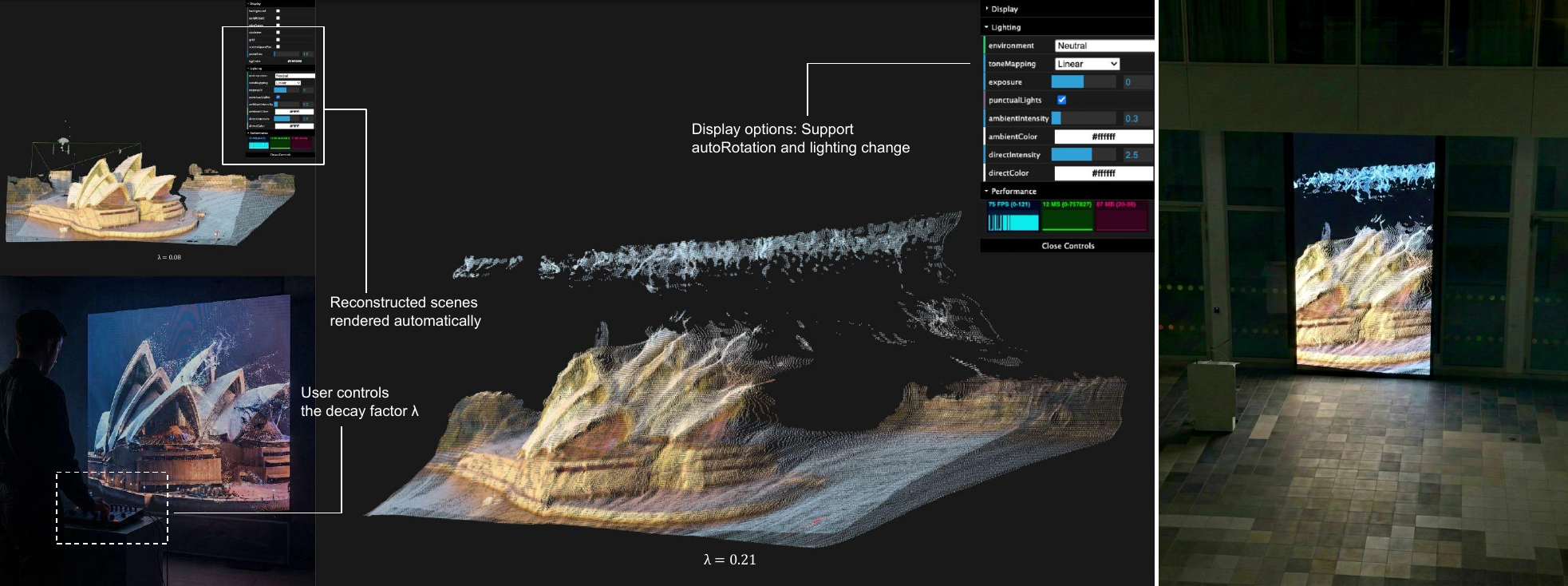}
    \caption{
    \textbf{Installation Design of \textit{Echoes of the Prior}.}
    Snapshot of the real-time feedback ($\lambda=0.21$). The viewport displays the immediate structural dissolution of the 3D scene as the user adjusts the decay factor. 
    }
    \label{fig:concept}
\end{figure}

\section{Related Work}

Positioning \textit{Echoes of the Prior} requires navigating the intersection of four distinct disciplines: computational aesthetics, artistic engagement with neurodegeneration, volumetric computer vision, and theoretical neuroscience.

\subsection{From Machine Vision to Machine Hallucination}
In the domain of AI art, early works focused on visualizing the \textit{constructive} capacity of neural networks. \textbf{DeepDream}~\cite{mordvintsev2015deepdream} and \textbf{Neural Style Transfer}~\cite{gatys2016image} exposed the internal representations of CNNs by maximizing feature activation, essentially asking the machine to ``see more'' of what it knows. Contemporary media artists like \textbf{Refik Anadol}~\cite{anadol2021machine} have further expanded this by treating massive datasets as fluid memories, visualizing the \textit{learning} process of the machine.

Our work inverts this paradigm. Instead of visualizing the \textit{genesis} of intelligence (learning), we visualize its \textit{entropy} (forgetting). We align with the tradition of \textbf{Glitch Art}~\cite{menkman2011glitch}, which views technological failure not as a bug, but as a revelation of the underlying medium. However, unlike traditional glitches that operate on the surface pixel level (e.g., datamoshing), our system performs \textbf{structural pathology}: we erode the internal cognitive weights and predictive priors themselves. This creates a form of ``Deep Glitch'' that emulates neurodegeneration rather than mere signal transmission error.

\subsection{Art, Memory, and Neurodegeneration}
Our work also situates itself within a broader artistic tradition engaging with memory loss and cognitive decline. 
One well-known precedent is the work of William Utermohlen, an artist diagnosed with Alzheimer's disease in 1995, whose self-portraits over subsequent years document a gradual shift from figurative realism toward abstraction, often interpreted as a first-person record of representational breakdown~\cite{crutch2001utermohlen}. 
Similarly, the late paintings of Willem de Kooning, produced during a period of cognitive decline, have sparked debate about whether their luminous simplicity reflects genuine artistic evolution or the influence of disease~\cite{espinel1996dekooning}. 
These cases demonstrate that neurodegeneration can produce visually compelling imagery while simultaneously raising complex ethical questions about authorship and artistic intent (see Sec.~\ref{sec:ethics}).

In the domain of interactive and new media art, Memo Akten's installation \textit{Learning to See}~\cite{akten2019learning} provides a related example: a neural network constrained by its training data hallucinates misreadings of live camera input, revealing how machine perception can become structurally distorted. 
Laurie Anderson and Hsin-Chien Huang's VR installation \textit{Chalkroom}~\cite{anderson2017chalkroom} similarly immerses viewers in rooms of dissolving words and narratives, evoking the fragility of linguistic memory.
Collaborative art -- science initiatives such as \textit{Created Out of Mind}~\cite{crutch2018createdoutofmind} at the Wellcome Collection have further brought together artists, neuroscientists, and people living with dementia to co-create works that reshape public understanding of these conditions.

Our contribution differs from these precedents in a key respect: rather than depicting memory loss through narrative, metaphor, or the artist's own declining hand, we engineer it directly within the cognitive architecture of the machine itself, creating what we term a ``Deep Glitch'' -- a structural pathology of the model's internal representations.

\subsection{Spatializing the Past: 3D Reconstruction as a Cognitive Medium}

While photography freezes a specific moment (a temporal slice), 3D reconstruction captures a specific environment (a spatial slice). We argue that \textbf{volumetric 3D representation} is intrinsically more akin to the phenomenology of biological memory than flat 2D imagery. Neuroscientific evidence suggests that memory encoding is deeply intertwined with spatial navigation (e.g., in the hippocampus)~\cite{okeefe1971hippocampus}. To remember is not just to view an image; it is to inhabit a space.

The evolution of 3D reconstruction, from explicit photogrammetry meshes to implicit \textbf{Neural Radiance Fields (NeRF)}~\cite{mildenhall2021nerf} and now \textbf{3D Gaussian Splatting (3DGS)}~\cite{kerbl20233dgs}, reflects a shift towards more organic representations of reality. Unlike polygonal meshes which are defined by rigid surfaces, volumetric approaches represent the world as a probabilistic field of density and color. 

In this work, we utilize 3DGS as our implementation carrier not merely for its real-time rendering capabilities, but for its unique aesthetic properties. Unlike generative video models that hallucinate temporal continuity, 3DGS reconstructs the \textit{spatial continuity} of a scene. By subjecting this volumetric representation to decay, we visualize the disintegration of the spatial context itself, simulating how a patient might lose their grip on the \textit{where} and \textit{how} of a memory, rather than just the \textit{what}.

\subsection{Computational Psychiatry and Predictive Coding}
Theoretically, our decay functions are grounded in the \textbf{Free Energy Principle}~\cite{friston2010free}, which posits that the brain minimizes surprise by constantly predicting sensory input based on internal priors. In this framework, psychopathology is often modeled as a failure of this predictive machinery~\cite{corlett2019hallucinations}.

Our implementation of the Sensory Decay ($\mathcal{G}_\text{sense}$) draws directly from the \textbf{Neural Gain Theory of Aging}~\cite{li2001aging}, which hypothesizes that cognitive decline is driven by a reduction in the signal-to-noise ratio (SNR) of neuromodulators (like dopamine) in the cortex. By injecting noise into the weights of our vision encoder, we create a functional analogy of this biological process~\cite{cichy2019deep}. Furthermore, our simulation of hallucinations aligns with models of \textbf{Computational Psychiatry}, which suggest that when sensory precision is low (as in our Spectral Amnesia module), the brain over-relies on top-down priors, leading to the perception of phantom stimuli~\cite{powers2017pavlovian}.

Complementing this, our Memory Decay ($\mathcal{F}_\text{mem}$) simulates the breakdown of top-down priors, grounded in the theory of \textbf{Attractor Dynamics}~\cite{hopfield1982neural}. In biological networks, memories are stored as stable states (attractors) within a high-dimensional manifold. Our geometric rotation of the latent space effectively destabilizes these attractors, simulating the phenomenology of \textbf{Semantic Dementia}~\cite{warrington1975selective}, a condition where the structural links between concepts erode. While the machine retains the raw sensory data, it loses the ``semantic glue'' that holds the geometry together, mirroring how patients with associative agnosia perceive form without meaning.

\section{Methodology: Simulating Cognitive Decay}

Modern foundational models~\cite{oquab2024dinov2, siméoni2025dinov3} do not store pixels; they store semantic concepts in a high-dimensional latent space. To simulate the phenomenology of forgetting, we do not merely degrade the image pixels; we perform surgical interventions on the cognitive architecture of the machine. 

Our approach creates a structural analogy between a state-of-the-art 3D reconstruction model~\cite{lin2025depth3recoveringvisual} based on feed-forward 3D Gaussian Splatting (FF-3DGS), and the Predictive Coding framework~\cite{friston2010free} of the human brain, synthesizing reality through the fusion of two distinct neural pathways: \textit{memory} (prior) and \textit{evidence} (sensation).

\subsection{The Silicon Proxy: Two Streams Mapping Mind to Code}
\textbf{\textit{Can a machine possess memory and cognition? If so, what does it look like?}}

The act of perceiving the world is a negotiation between what our eyes see -- the evidence from \textit{Sensory Cortex}, and what our brains expect -- what neuroscience terms the \textit{Engram}, the physical trace of memory. We draw a functional analogy between this biological duality and the specific components of our computational architecture (see Sec.~\ref{sec:limits} for a discussion of the boundaries of this mapping):

\subsubsection{The Sensory Stream: Processor Weights} 
\label{sec:sensory_define}

Drawing a functional analogy to the biological retina and V1 cortex, our model employs a shallow CNN to process raw pixel inputs, capturing the texture and local geometry of the \textit{Now}.
This sensory signal will serve as a reinforcement to the memory, and together with memory, will be injected into a fusion block for spatial perception and reconstruction~\cite{ranftl2021vision}. 

\subsubsection{The Memory Stream: Semantic Features}
\label{sec:memory_define} 
In an analogy to how biological memory stores concepts rather than raw data, our system utilizes vision foundational models to encode high-dimensional semantic priors.

\begin{figure}[H]
    \centering
    \includegraphics[width=0.95\linewidth]{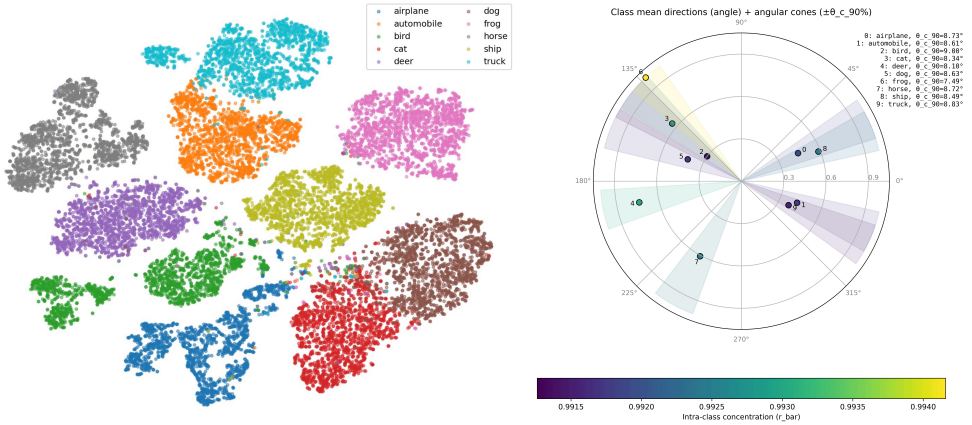}
    \caption{
        t-SNE visualization of the DINOv2 latent space. \textbf{(Left)} Semantic concepts cluster into distinct manifolds, forming the structured \textit{healthy} memory baseline. 
        \textbf{(Right)} Cosine similarity confirms that intra-class features are more closely aligned than inter-class features, validating the semantic coherence of the latent space.
    }
    \label{fig:PriorShape}
\end{figure}

Specifically, we use \textbf{\textit{DINOv2}}~\cite{oquab2024dinov2}, a state-of-the-art visual understanding model as the \textit{prior} producer in our feed-forward 3D reconstruction model. 
To give an intuitive sense of this prior, we visualize its latent representation space. 
As shown in Fig.~\ref{fig:PriorShape} on the left, in this high-dimensional (1024D) latent space, for instance, visual features of the same semantic concept cluster together and form a distinct manifold, diverging significantly from others in terms of geometric similarity metrics. 
We employ t-SNE and PCA to project these 1024D latent vectors into 2D space for better visualization. 
These latent representations serve as the \textbf{\textit{healthy baseline}}: the long-term memory of a healthy silicon brain -- structured, dense, and semantically coherent.

In a healthy state, perception is the successful integration of these two streams. Our installation visualizes what happens when these streams decay asynchronously.

\subsection{The Mechanics of Forgetting: Two Decay Functions Simulating Biological Entropy}
Human aging is characterized by two distinct processes: sensory degradation and cognitive decline. 
To simulate both processes, we introduce two distinct entropy operators, $\mathcal{G}_\text{sense}$ and $\mathcal{F}_\text{mem}$, grounded in the \textit{Neural Gain Theory of Aging}~\cite{li2001aging}, which posits that both types of degradation are driven by a reduction in the signal-to-noise ratio within specific neuromodulatory systems.

\subsubsection{Sensory Decay: The Erosion of Visual Primitives $\mathcal{G}_\text{sense}$}
\label{sec:sensory_decay}

While the semantic stream deals with concepts, the sensory stream deals with \textit{visual primitives} -- edges, textures, and local gradients. A naive simulation of sensory decay would simply degrade the input signal (i.e., pixel-level corruptions), simulating external visual impairments. However, biological aging is a neurological process where the processing circuitry itself degrades. 

To simulate this, we perform a perturbation on the CNN weights that acts as the machine's primary visual cortex (as described in Sec.~\ref{sec:sensory_define}). We implement a compound decay function $\mathcal{G}_\text{sense}$ that creates a diseased clone of the sensory encoder, modeling three distinct biological failures:

\begin{figure}
    \centering
    \includegraphics[width=0.98\linewidth]{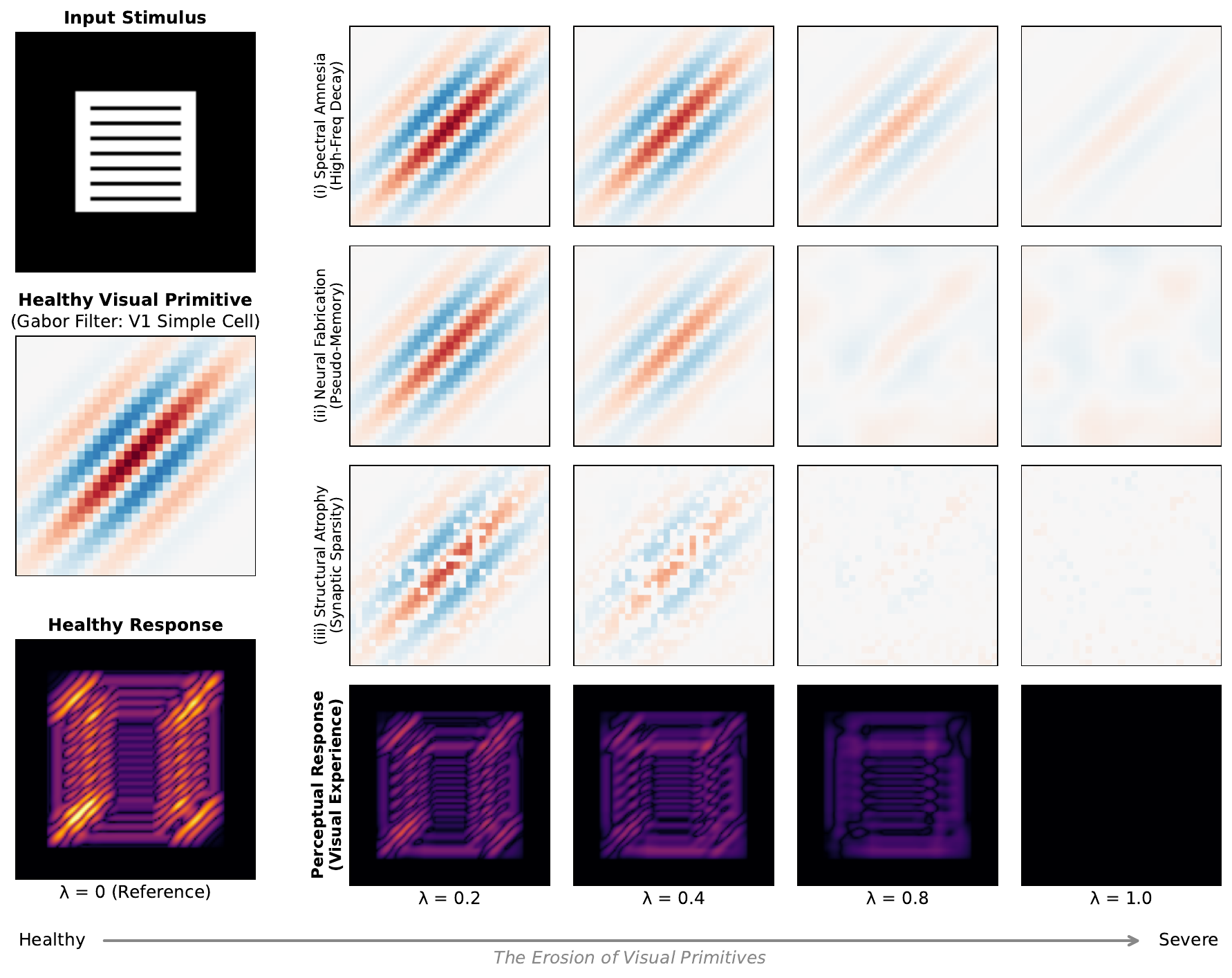}
    \caption{
        \textbf{Simulating Sensory Decay $\mathcal{G}_\text{sense}$.} 
        \textbf{(Left) Healthy Baseline ($\lambda=0$): } Response of a standard Gabor filter (V1 primitive) to a sharp stimulus.
        \textbf{(Right) Pathology: } As $\lambda$ increases, the filter degrades via three mechanisms: \textbf{(i) Spectral Amnesia} (loss of high-frequency acuity); \textbf{(ii) Neural Fabrication} (injection of coherent noise/hallucinations); and \textbf{(iii) Structural Atrophy} (synaptic sparsity leading to signal loss).
    }
    \label{fig:SensoryDecay}
\end{figure}

\noindent \textbf{(i) Spectral Amnesia (High-Frequency Domain Decay):} 
Biological aging often begins with the loss of high-frequency acuity~\cite{owsley1983contrast}. 
To simulate this, we transform the convolutional kernels into the frequency domain via Fast Fourier Transform (FFT) and then apply a radial mask that selectively dampens high-frequency components relative to the decay factor $\lambda$ \footnote{The decay factor in $\mathcal{G}_\text{sense}$ can be distinct from the one in memory decay $\mathcal{F}_\text{mem}$ for asynchronous control, and we keep the same symbols here to avoid complicating the notation.}. 
This operation is grounded in the functional architecture of the primary visual cortex (V1), where simple cells act as spatiotemporal frequency filters~\cite{de1982spatial}. By dampening high frequencies in the kernel weights directly, rather than the input image, we simulate the degradation of the neurons' receptive fields themselves, rendering them incapable of resonating with fine details.
\textbf{The Effect:} As shown in Fig.~\ref{fig:SensoryDecay}, the machine does not see a blurry image; it simply forgets how to detect sharpness. The neural filters responsible for recognizing crisp edges and fine textures (modeled as Gabor-like simple cells~\cite{hubel1962receptive}) are chemically suppressed, simulating the onset of cortical cataracts.

\noindent \textbf{(ii) Neural Fabrication (Pseudo-Memory Injection):} 
The brain abhors a sensory vacuum. When visual inputs degrade, the visual cortex often generates spontaneous activity to fill the void (e.g., Charles Bonnet Syndrome~\cite{reichert2013charles}). We model this by injecting spatially smoothed Gaussian noise into the weights. 
This approximates the ``correlated noise'' observed in neural circuits when inhibitory control weakens~\cite{shadlen1998variable}. 
Unlike white noise, this smooth noise creates coherent but fictitious patterns in the convolution filters. \textbf{The Effect:} This causes the machine to \textit{hallucinate} textures that do not exist, overlaying the real world with a dream-like, phantom grain.

\noindent \textbf{(iii) Structural Atrophy (Synaptic Sparsity):} Finally, we simulate the physical death of neurons. We apply a stochastic mask to the \textit{Weight} matrix, zeroing out connections based on $\lambda$. Simultaneously, the \textit{Bias} terms in the convolutional layers are decayed towards zero, representing a loss of neural ``confidence'' or activation potential. \textbf{The Effect:} This creates a \textit{sparse} perception and the image becomes \textit{ghostly}. As the machine loses the confidence to assert the existence of matter, the reconstructed 3D world becomes translucent and ethereal, creating a visual metaphor for a consciousness that is slowly detaching from physical reality.

\subsubsection{Memory Decay: Manifold Distortion and Cognitive Entropy $\mathcal{F}_\text{mem}$}
\label{sec:memory_decay}

\begin{figure}[t]
    \centering
    \includegraphics[width=0.99\linewidth]{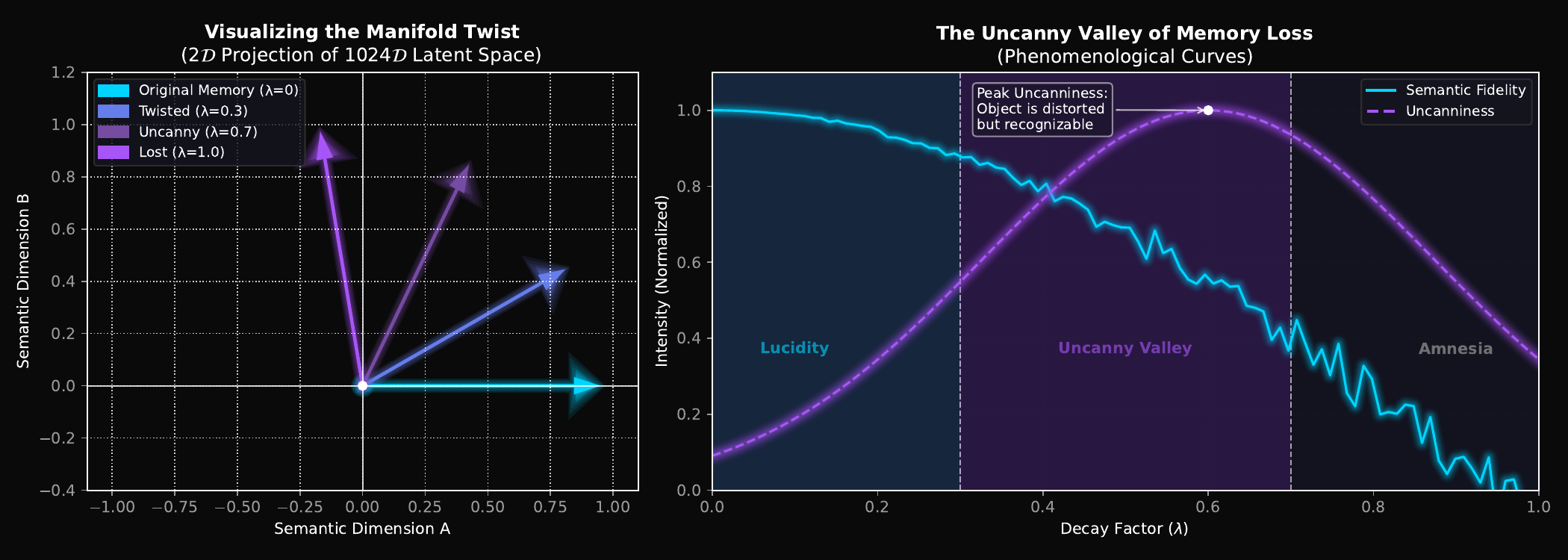}
    \caption{ 
        \textbf{Topology of Memory Decay.}
        \textbf{(a) Manifold Twist: }Visualization of the orthogonal rotation $\mathbf{R}(\lambda)$. As $\lambda$ increases (blue $\to$ purple), the feature vector rotates away from its semantic alignment while preserving its norm.
        \textbf{(b) Computational Uncanny Valley: } A computationally defined model of visual \textit{Uncanniness} (purple), defined as the product of structural integrity (norm) and semantic distortion ($1 - \text{cosine similarity}$).
        The peak ($0.3 < \lambda < 0.7$) corresponds to the zone where the representation is structurally strong yet semantically twisted. Perceptual research predicts that such feature-inconsistent stimuli are maximally disturbing~\cite{seyama2007uncanny, moore2012bayesian}; formal viewer validation remains future work.
    }
    \label{fig:MemoryDecay}
\end{figure}

Biological forgetting is often a structural collapse of meaning, where the world does not just fade into black but turns disturbingly wrong. To simulate the phenomenology of \textit{Semantic Dementia}~\cite{warrington1975selective} and \textit{Associative Agnosia}~\cite{farah2004visual}, we model the machine memory as a high-dimensional vector field (as introduced in Sec~\ref{sec:memory_define}), where the semantic identity of an object is defined mainly by the direction of its \textbf{feature vector} $\mathbf{z}$ ~\cite{wang2020understanding}, and introduce a \textit{geometric perturbation} within the high-dimensional memory space. 

\textbf{The Distortion:} We model \textit{cognitive decline} as a force that \textbf{\textit{twists}} the memory manifold.  
Using a continuous orthogonal rotation $\mathbf{R}(\lambda)$, we misalign semantic vectors while preserving their norm: 
\begin{equation}
    \mathbf{R}(\lambda) = (1 - \lambda) \cdot \mathbf{I} + \lambda \cdot \mathbf{Q}, 
\end{equation}
where $\mathbf{Q}$ is a random orthogonal matrix generated via QR decomposition~\cite{mezzadri2007generate}, $\mathbf{I}$ is the identity matrix. As the decay factor $\lambda$ increases, the semantic feature vectors $\mathbf{z}$ are smoothly rotated away from their ground truth alignment. 
\begin{equation}
    \mathbf{z}_\text{rot} = \mathbf{z} \cdot \mathbf{R}(\lambda)^T. 
\end{equation}

This simulates a brain state where neural firing rates (signal strength) remain high, but the semantic connectivity (signal direction) is twisted -- suffering from ``cognitive confusion''. The machine continues to reconstruct complex, high-confidence geometries, but because the vectors are misaligned with the original priors, familiar objects are rendered as \textit{hallucinated} topological variants. For instance, as shown in Fig.~\ref{fig:teaser}, a construction might be reconstructed with similar texture but the geometry of a liquid.

\textbf{The Hybrid Entropy:} Furthermore, to capture the full spectrum of memory decay, we combine this geometric twisting with a secondary entropy term, reflecting the irreversible loss of information:
\begin{equation}
\label{eq:memo_decay}
    \mathcal{F}_{mem}(\mathbf{z}, \lambda, \gamma) = \underbrace{(1 - \gamma)\cdot \mathbf{z}_\text{rot}}_{\text{Distortion}} + \underbrace{\gamma \cdot \mathcal{N}(\mathbf{z})}_{\text{Entropy}}
\end{equation}
\noindent where $\mathbf{z}, \mathbf{z}_\text{rot} \in \mathbb{R}^D$ are feature vectors (here $D=1024$), function $\mathcal{N}(\mathbf{z})$ generates a stochastic noise vector, $\mathcal{N}(\mathbf{z}) \approx \epsilon \cdot \|\mathbf{z}\|$, where $\epsilon \sim \mathcal{N}(0, I)$.   
By fusing manifold rotation ($1 - \gamma = 0.7$) with stochastic noise ($\gamma = 0.3$), the system visualizes a specific trajectory of forgetting: as shown in Fig.~\ref{fig:MemoryDecay}, memory first becomes confused (twisted geometry) before it eventually becomes lost (dissolved form). 
This algorithmic choice allows us to visualize what we term a \textit{computational uncanny valley} of memory loss -- a regime ($0.3 < \lambda < 0.7$) where the product of preserved structural integrity and semantic distortion peaks. We note that this ``valley'' is defined computationally, not through measured viewer responses; however, it is consistent with perceptual research showing that stimuli exhibiting high structural fidelity but semantic inconsistency produce maximal discomfort~\cite{seyama2007uncanny, moore2012bayesian} -- an effect Freud~\cite{freud1919uncanny} described as \textit{das Unheimliche}, the disturbing quality of the familiar-yet-wrong.
This zone conceptually mirrors the phenomenology of Associative Agnosia~\cite{farah2004visual} and Semantic Dementia~\cite{warrington1975selective}, where the patient perceives form without meaning.

\subsection{The Selective Forgetting: Object-Oriented Amnesia}

While global decay simulates a systemic cognitive decline (e.g., dementia), human forgetting is often highly selective~\cite{warrington1975selective}. We frequently lose the semantic grasp of specific entities -- a face, a name, or an object -- while the surrounding reality remains intact. To simulate this \textit{dissociative amnesia}, we introduce a targeted decay mechanism that allows the system to surgically erode the memory of specific semantic categories.

\subsubsection{Semantic Grounding and Masking}
\begin{figure}[h]
    \centering
    \includegraphics[width=0.99\linewidth]{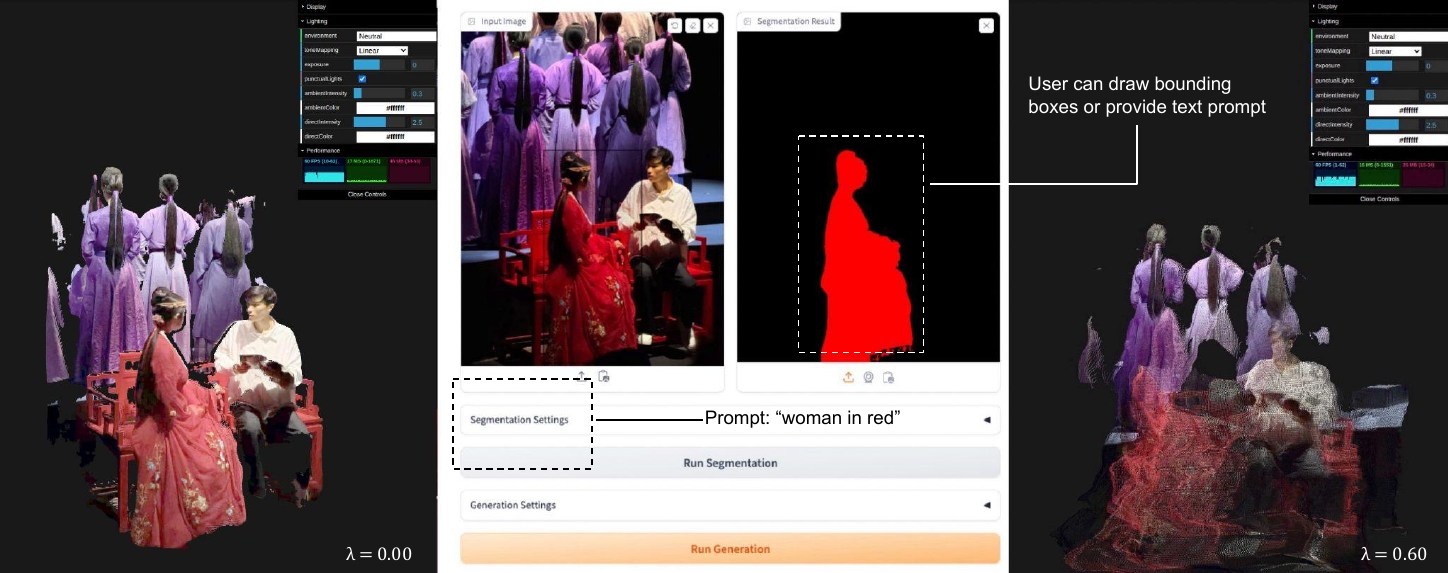}
    \caption{
        \textbf{Visualizing Selective Forgetting.}
        Targeted entropy injection using a text prompt ("woman in red"). \textbf{(Left)} Original reconstruction. \textbf{(Right)} Decayed result ($\lambda=0.6$). The targeted entity undergoes topological dissolution, effectively ghosting out of reality while the surrounding context remains intact.
    }
    \label{fig:SelectDecay}
\end{figure}

We employ a two-stage pipeline to isolate the \textit{memory trace} of a specific object. First, we utilize an open-vocabulary detection model, \textbf{\textit{Grounded-SAM}}~\cite{ren2024grounded}, to generate a pixel-level binary mask $M_{pixel} \in \{0, 1\}$ based on a textual prompt (e.g., ``woman in red''). This mask represents the object's footprint in the visual field. As shown in Fig.~\ref{fig:SelectDecay}, users can obtain the mask for specific entities by providing a text prompt or drawing bounding boxes. A naive approach would be to apply the mask to the input image directly (pixel-level inpainting), but this would result in a superficial ``black hole'' rather than a cognitive loss. Instead, we perform the intervention within the latent feature space of the memory.

\subsubsection{Latent Space Surgery}
To extract the prior about the input images, the vision foundation model~\cite{oquab2024dinov2} first processes images as a sequence of patch tokens. To align with the spatial shape of these patch tokens, we down-sample the pixel mask $M_\text{pixel}$ to create a discretized token-wise mask $M_\text{token} \in \{0, 1\}$. 

We then modify the forward pass of our system: we perform decay function Eq.~(\ref{eq:memo_decay}) only into the tokens corresponding to the target object:
\begin{equation}
    \mathbf{z}' = M_\text{token} \cdot \mathcal{F}_{mem}(\mathbf{z}, \lambda, \gamma) + (1 - M_\text{token}) \cdot \mathbf{z},
\end{equation}
where $\mathbf{z}$ is the original clear memory.

By perturbing only the latent tokens of the target, we achieve a form of \textit{selective agnosia}. As shown in Fig.~\ref{fig:SelectDecay} on the right, the reconstructed 3D point cloud of the surrounding environment (e.g., other actors) remains structurally sound. However, the targeted entity (e.g., \textit{``woman in red''}) begins to exhibit a topological failure: its geometry melts, its texture turns into noise, and it effectively ghosts out of physical existence. 
This visualizes the psychological state where a patient may perceive the context of a room perfectly but fails to resolve the identity or form of a specific object within it -- a computational emulation of the brain's refusal to process a traumatic memory. We give more discussion in Sec~\ref{sec:artifact_select}. 

\section{Real-time Installation and Implementation Details}

\begin{figure}[t]
    \centering
    \includegraphics[width=0.99\linewidth]{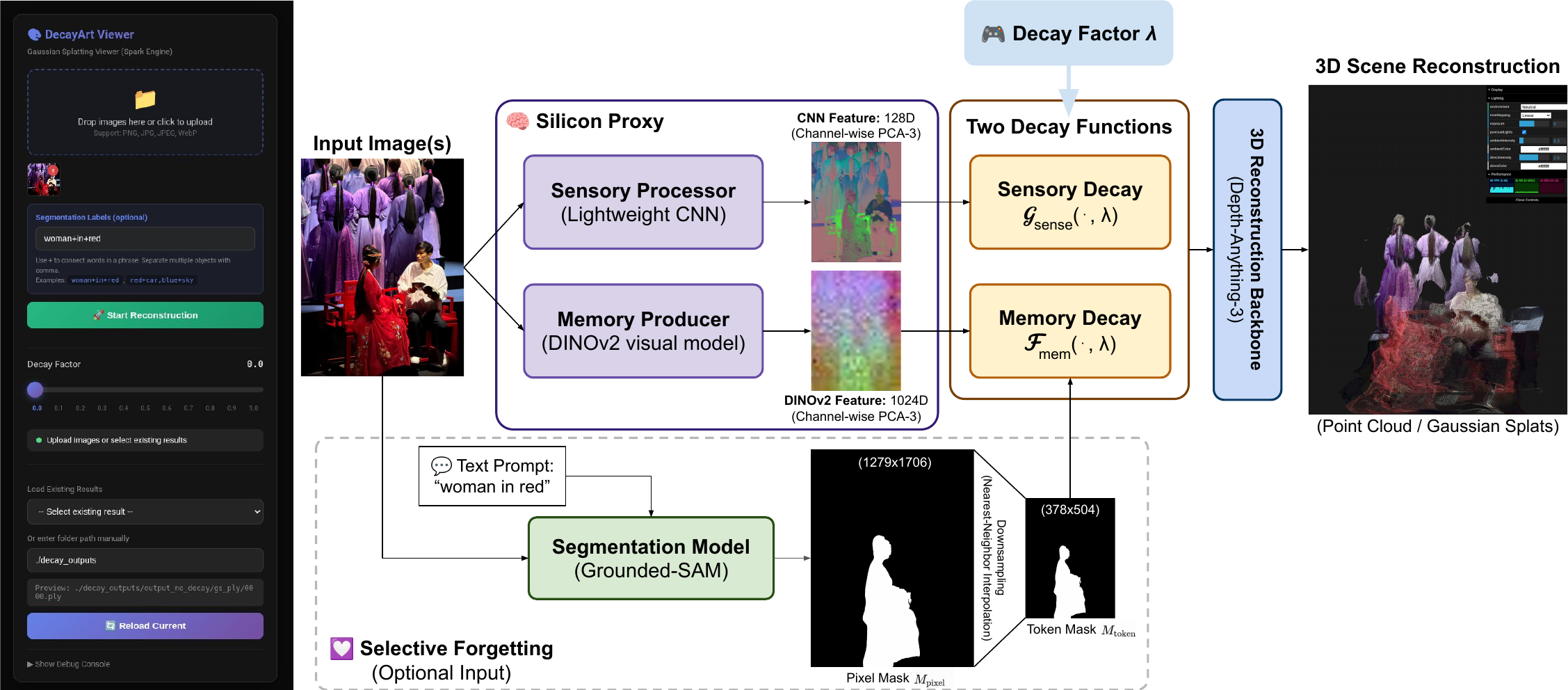}
    \caption{The System Workflow of \textit{Echoes of the Prior}.
    }
    \label{fig:pipeline}
\end{figure}

\subsection{User Interface and Integration Formula} 

We designed \textit{Echoes of the Prior} as a real-time interactive installation. 
The core of the installation is a closed-loop reconstruction system that processes visual stimuli into decaying 3D memories. 

As shown in Fig.~\ref{fig:pipeline}, the system accepts visual input (either from a live camera feed capturing the viewer's environment or pre-loaded image dataset) and processes it through a pre-trained 3D reconstruction backbone, \textbf{\textit{Depth Anything 3}}~\cite{lin2025depth3recoveringvisual}, for memory construction. 
Unlike standard inference, the user-defined decay factor $\lambda$ \textbf{intercepts} this reconstruction process by two parallel decay functions: \textbf{The Sensory Decay Path (Top)}, which acts as a spectral filter (see Sec.~\ref{sec:sensory_decay}); \textbf{The Memory Decay Path (Bottom)}, which optionally integrates a \textbf{Selective Forgetting} module.
This module utilizes \textbf{\textit{Grounded-SAM}}~\cite{ren2024grounded} to generate semantic masks based on user prompts. Entropy is \textbf{surgically injected} only into the tokens representing specific objects, disrupting their structural integrity while preserving the context. 
Finally, these corrupted latents from both paths are fused together and then decoded by the 3D reconstruction backbone into a fractured 3D scene represented by Gaussian Splats or point cloud, which is rendered in real-time via a WebGL-based viewer to provide immediate visual feedback of the decay. 
The system supports 60FPS interaction: 
as the user slides the fader, the world melts instantly, creating a visceral connection between their action and the visual destruction.

\begin{figure}[b]
    \centering
    \includegraphics[width=0.99\linewidth]{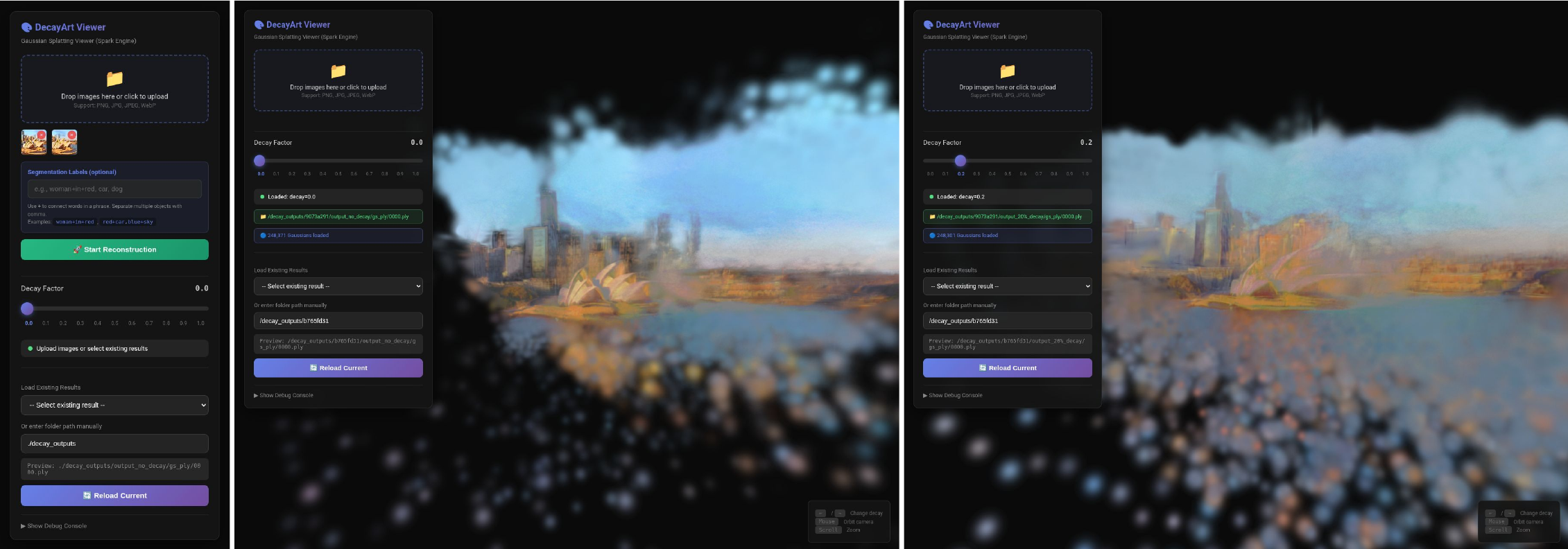}
    \caption{
        \textbf{The Interactive Interface.} The web-based GUI enables real-time interaction with the decay process.
    }
    \label{fig:interface}
\end{figure}
 
Designed with minimalist semantics, the interface shown in Fig.~\ref{fig:interface} is centered on a single slider ($\lambda \in [0, 1]$). This transforms an abstract algorithm into a proprioceptive experience: as the user physically drags the fader, they witness the immediate entropic dissolution of the scene. 
Unlike static video loops, the rendered scene supports full orbit controls. Users can rotate and zoom into the decaying structures, inspecting the glitch artifacts up close.

\subsection{Image Acquisition Strategy}

The system supports dual inputs: a \textbf{Curated Archive} of semantic archetypes for immediate engagement, and a \textbf{Personal Injection} feature allowing users to upload personal photos or video, transforming the installation into a digital elegy.

\begin{figure}[H]
    \centering
    \includegraphics[width=0.99\linewidth]{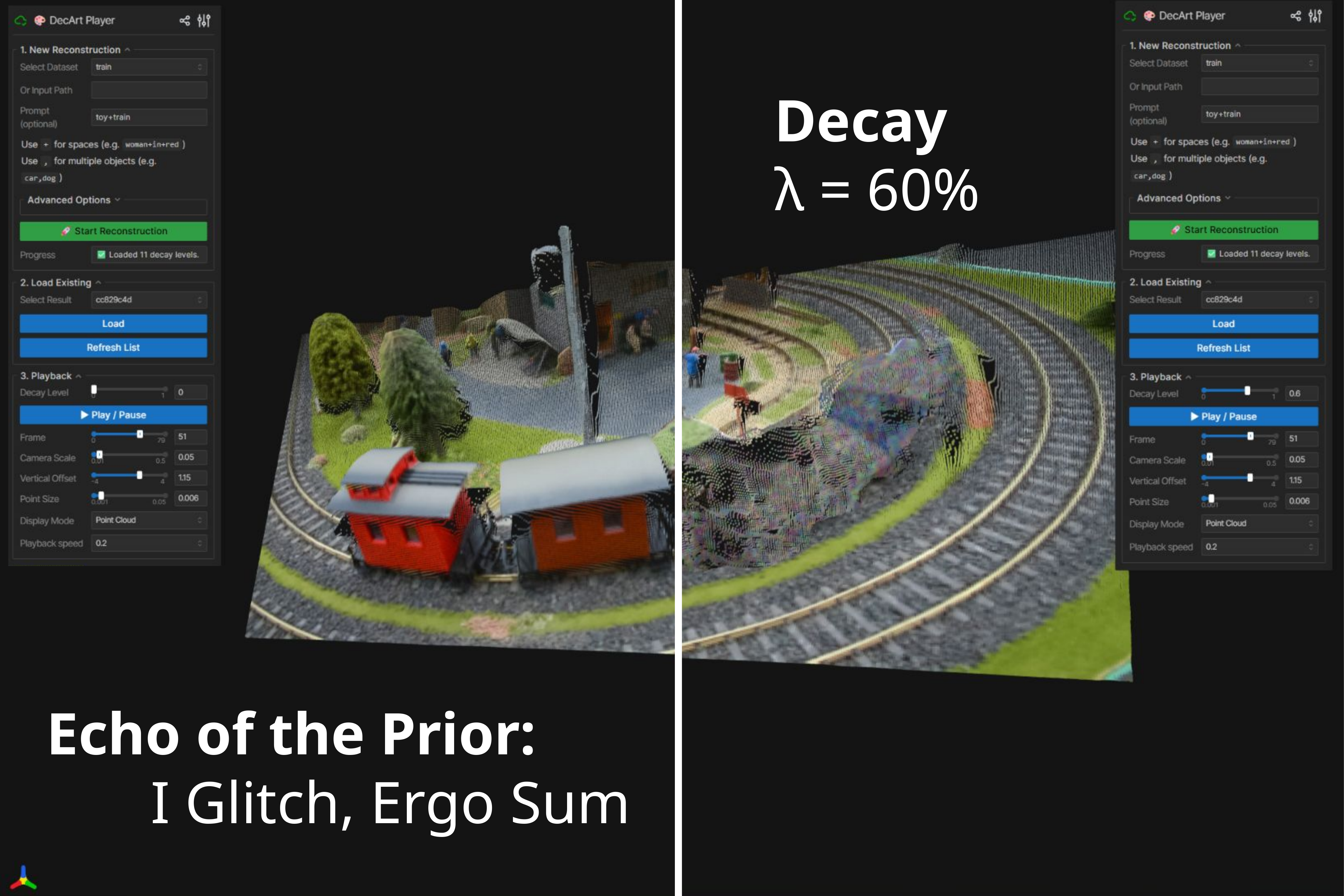}
    \caption{
        \textbf{``I Glitch, Ergo Sum'' (I glitch, therefore I am).} 
        A visual manifesto generated by our real-time streaming system \textit{\textbf{DecArt}}. 
    }
    \label{fig:streaming}
\end{figure}

While the paper figures illustrate the core algorithm, we developed an advanced \textit{\textbf{Interactive Streaming Interface}} (shown in the demo video, a representative frame shown in Fig.~\ref{fig:streaming}) to facilitate real-time experimentation. This system integrates Grounded-SAM for prompt-based masking and implements a batch pre-computation pipeline, allowing users to seamlessly visualize the continuous decay process in 4D space without inference latency.

\subsection{The Experience Journey}
\label{sec:experience}

We describe the intended audience encounter to articulate the experiential arc of the installation.

\noindent \textbf{Encounter and Orientation.}
The viewer enters a darkened space and approaches a large display showing a photorealistic 3D scene, with a tablet serving as the control console. At $\lambda = 0$, the scene appears pristine -- a faithful reconstruction that can be orbited freely. The scene may be drawn from a \textit{Curated Archive} or, in a more intimate \textit{Bring Your Own Memory} mode, from the viewer's own personal video uploaded via AirDrop or QR code.

\noindent \textbf{Two Modes of Interaction.}
The viewer controls the decay through two complementary modes. In \textit{Global Decay}, a slider governs $\lambda$ directly: as it increases, edges soften, textures hallucinate, and geometry warps -- progressing from subtle shimmer ($\lambda \approx 0.1$) through an unsettling intermediate zone of topological distortion ($0.3 < \lambda < 0.7$) to total semantic dissolution ($\lambda > 0.8$). The slider is bidirectional, letting the viewer retreat from oblivion at any point. In \textit{Selective Forgetting}, the viewer names or taps a specific element (e.g., ``the woman in red''); the system segments the target via Grounded-SAM~\cite{ren2024grounded} and restricts the entropy injection to its latent tokens alone. The result is a world perfectly remembered except for one conspicuous absence -- a ghost-shaped void. When operating on a personal video, selecting and dissolving a specific figure becomes a deeply personal gesture.

\noindent \textbf{Sonic Dimension.}
For the exhibition setting, we envision a generative soundscape that co-evolves with the visual decay: spectral filtering to strip harmonics, granular synthesis to introduce stochastic micro-events, and amplitude attenuation toward silence -- all parametrically coupled to $\lambda$, extending the experience into a coherent multimodal dissolution.

\noindent \textbf{Aesthetic Character: The Deep Glitch.}
The resulting visual language -- which we term \textit{semantic liquefaction} -- is distinct from both traditional Glitch Art (broken data) and generative AI aesthetics (fabricated coherence). Because the decay operates on cognitive representations rather than pixels, objects retain their texture while their geometry undergoes topological metamorphosis: a building does not pixelate but \textit{melts}, retaining just enough identity to be recognized as a ruin of itself. The volumetric medium further enriches this effect -- orbiting the scene reveals that corruption is non-uniform, with some vantage points exposing catastrophic voids behind near-coherent facades, giving the work a sculptural rather than pictorial character.

\subsection{Negative Justification: The Ontology of the Shell}
In the development of \textit{Echoes of the Prior}, we evaluated two distinct paradigms for 3D scene reconstruction: (1) generative multi-view diffusion, represented by MIDI-3D~\cite{huang2025midi}, and (2) monocular depth reconstruction, represented by Depth Anything V3 (DA3)~\cite{lin2025depth3recoveringvisual}.

\begin{figure}[H]
    \centering
    \includegraphics[width=0.99\linewidth]{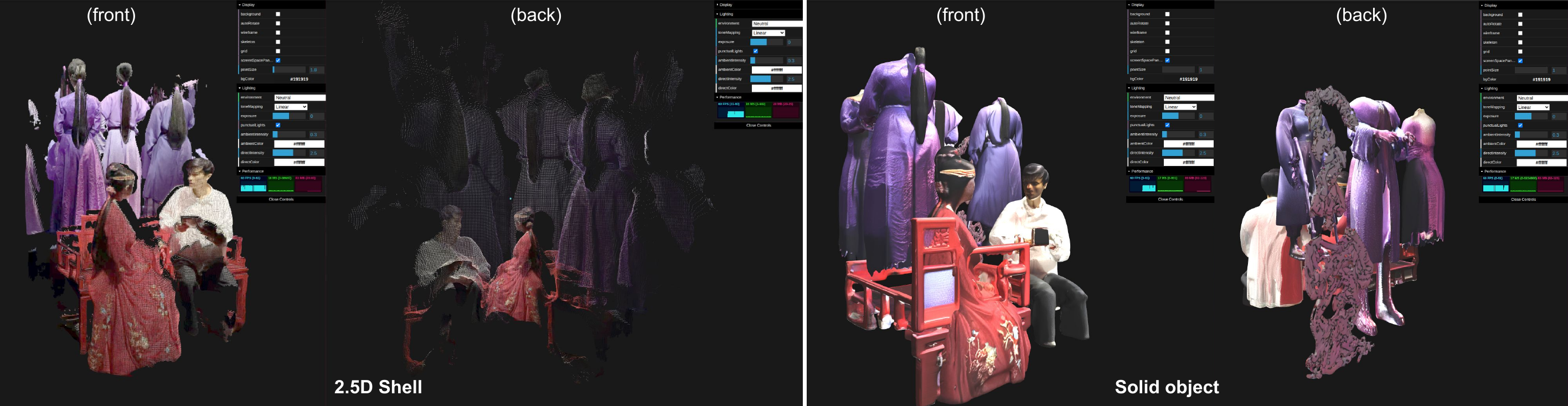}
    \caption{\textbf{Memory as a Potemkin Village: Hallucination vs. Reconstruction.} While 3D generative models \textbf{(Right) }fabricate a complete, watertight object by inventing unseen details, our system \textbf{(Left)} preserves the epistemological honesty of the input. It projects a thin, fragile film, like a \textit{facade} of memory, that disintegrates when the viewer attempts to look behind the veil. This hollowness is not a technical failure, but a deliberate aesthetic feature representing the void of forgotten context.}
    \label{fig:shell}
\end{figure}

While generative models offer watertight meshes, as shown in Fig.~\ref{fig:shell}, we find such completeness antithetical to memory. Memory is a surface, not a solid~\cite{merleau1945phenomenology}.
We therefore choose DA3's monocular reconstruction to preserve the \textit{Potemkin village effect}: a fragile film that reveals its void upon rotation, metaphorically representing the emptiness of forgotten contexts.

\section{Discussion}

\subsection{Design Insights: Glitcho, Ergo Sum} 
\label{sec:name}
The homophone \textit{DecArt} invites a reflection on \textit{Cartesian doubt} in the age of silicon. Descartes famously stripped away all sensory perception as potentially deceptive, arriving at the only undeniable truth: \textit{Cogito, Ergo Sum} (I think, therefore I am). 
Our installation poses a parallel question for the machine: when the sensory weights rot and the semantic priors twist, does the ``ghost in the shell'' still exist? Or, in our case: \textit{Glitcho, Ergo Sum} (I glitch, therefore I am).

This suggests a profound irony: a perfectly optimized neural network is merely a sterile photocopier -- accurate, yet alien. The ``humanity'' of the machine emerges not in its computation, but in its poetic failure. When the AI struggles to reconstruct reality from corrupted memory, it ceases to be a tool and becomes a mirror, forcing the audience to confront the inevitability of their own cognitive entropy. 
During early presentations of the work, viewers were invited to interact with the system by adjusting the decay parameter and observing the progressive dissolution of the reconstructed scene. 
Viewers spent extended periods exploring intermediate decay levels of the system, where objects remained partially recognizable yet semantically distorted. 
A recurring reaction -- especially among viewers familiar with computer vision or 3D reconstruction -- was surprise at this reframing: 3D reconstruction is typically evaluated purely in terms of fidelity, whereas the intentional and structured degradation of reconstruction quality was perceived as opening an unexpected aesthetic and conceptual design space. 
These informal observations suggest that the installation can prompt audiences to reinterpret reconstruction models not only as tools for accurate recovery, but also as media for exploring perceptual and cognitive fragility. 
These observations are qualitative and are not intended as a formal user study. 
Ultimately, we argue that generative AI can be more than an engine for productivity; it can be a medium for \textit{radical empathy}. 

\subsection{Ethical Dimensions: On Aestheticizing Neurodegeneration}
\label{sec:ethics}
Any artistic engagement with neurodegeneration must confront an ethical tension: the risk that aestheticizing cognitive decline reduces lived suffering to visual spectacle~\cite{sontag2003regarding}. Cultural representations of dementia -- as ``living death'' or ``loss of self'' -- can reinforce stigma and strip dignity from those living with these conditions~\cite{zeilig2013dementia}.
We acknowledge this tension directly. Our intent is not to aestheticize suffering but to cultivate \textit{empathetic understanding} -- to let the viewer briefly inhabit a perceptual world that is structurally disorienting, fostering compassion rather than voyeurism. In this, we align with person-centered frameworks~\cite{kitwood1997dementia} that emphasize the retained capacities and subjective experience of people with dementia, rather than framing the condition solely as loss~\cite{kontos2005embodied}.
We also acknowledge that the current project was developed without direct engagement with communities affected by neurodegeneration. Future iterations should involve collaboration with patients, caregivers, and clinicians to ensure the work serves as a bridge toward understanding rather than an appropriation of experience~\cite{basting2009forget}.

Early conversations around the work also extended beyond the immediate visual experience. 
Viewers from humanities and literary backgrounds often interpreted the installation as a provocation about the limits of machine subjectivity. 
If contemporary generative systems can simulate the appearance of memory, confusion, or loss without possessing first-person sensation, in what sense can such outputs be understood as artistic expression rather than secondary descriptions of human experience? 
Rather than resolving this question, the project intentionally leaves it open, positioning the installation as a space for reflection on the relationship between artificial generation, perception, and authorship.

\section{Limitations and Future Work}

\subsection{The Boundaries of the Silicon Proxy}
\label{sec:limits}

We wish to be explicit about the epistemological status of our work. The mappings we draw between computational components and biological processes -- CNN weights to V1 simple cells, DINOv2 features to memory engrams -- are \textit{productive artistic analogies}, not claims of mechanistic equivalence. Deep neural networks can predict neural responses in the visual cortex~\cite{yamins2016using}, but this correspondence is statistical rather than mechanistic: architecturally diverse models achieve similar predictive performance~\cite{storrs2021diverse}, and DNNs diverge from biological vision in fundamental ways~\cite{bowers2023deep, lindsay2021convolutional}. The mapping between any specific architecture and the brain remains underdetermined~\cite{jonas2017neuroscientist}.

Our system is therefore best understood as a \textit{speculative computational phenomenology}~\cite{varela1996neurophenomenology}: it uses the structural analogy between neural network pathology and cognitive decay as an artistic lens~\cite{cichy2019deep}, aligned with the tradition of using computational systems as ``metapictures''~\cite{offert2021perceptual, zylinska2020ai} rather than literal scientific models.
Moreover, while we successfully simulate the visual phenomenology of memory loss (the \textit{look} of forgetting), we cannot simulate the affective sensation (the \textit{feeling} of sadness). The machine does not miss the objects it forgets; it simply processes noise. This highlights the \textit{Hard Problem of Consciousness}~\cite{chalmers1995facing}. We do not claim to model how the brain actually degenerates; rather, we ask: \textit{if} a machine's cognition were organized analogously to ours, \textit{what would its forgetting look like?} Future work could integrate \textit{Affective Computing} modules to modulate the decay based on a simulated \textit{emotional state} of the AI.

\subsection{Simplified Cognitive Pathway}

Another limitation of our framework lies in the simplified modeling of the cognitive pathway between perception and long-term memory.
In cognitive psychology, the formation of stable long-term representations typically involves intermediate processes -- attention, working memory~\cite{baddeley1992working}, and information integration -- that mediate between sensory input and memory consolidation~\cite{atkinson1968human}.
Our current implementation bypasses these stages, modeling forgetting solely as a decay in sensory processing and semantic memory priors.

While this abstraction yields a clear and controllable computational framework, it omits mechanisms that may play a significant role in the phenomenology of memory distortion.
For example, failures in attentional selection or disruptions in working memory could produce qualitatively different forms of altered perception.
Future work may explore whether introducing such intermediate stages -- attentional filtering, capacity-limited buffers -- could generate new visual distortion aesthetics, further expanding the design space of computational phenomenology.

\subsection{Artifacts in Selective Forgetting: The Impossibility of Clean Forgetting}
\label{sec:artifact_select}
Our current implementation of object-aware forgetting relies on mapping pixel-level segmentation masks to the lower-resolution latent space of the memory. The necessary down-sampling and interpolation of the binary masks result in imperfect alignment at object boundaries, leading to a halo effect where the decay noise spills over into adjacent pixels. While technically a precision error, this limitation may align with the project's conceptual framework. It visualizes the impossibility of isolating a single memory trace without affecting its connected reality -- the impossibility of \textit{clean forgetting}. Just as a fading memory often blurs the details of its spatial context, our installation's inability to perform a clean cut reflects the entangled nature of visual perception.

\subsection{Future Work: Towards a Romantic Neuromorphic Art}
While standard Artificial Neural Networks (ANNs) constitute the backbone of current AGI, they represent a static ``snapshot'' of thought. To simulate \textit{living} memory, we aim to transition to Spiking Neural Networks (SNNs)~\cite{maass1997networks}. Though currently confined to neuromorphic robotics, we repurpose this event-driven technology to visualize the \textit{silencing of time}, where forgetting is not just noise, but the cessation of firing.

\begin{acks}
Andreas Geiger is a member of the Machine Learning Cluster of Excellence, EXC 2064/1, project number 390727645. The authors gratefully acknowledge the ML Cloud and the Tübingen AI Center for providing the computational resources and facilities that supported this work. We thank Prof. Jakob H. Macke for his thoughtful comments on the cognitive and neuroscientific aspects of the manuscript, and Dr. Canmei Xu (KU Leuven) for her input from a cognitive psychology perspective.
\end{acks}

\bibliographystyle{ACM-Reference-Format}
\bibliography{acmart}

\end{document}